\documentclass[10pt,twocolumn,letterpaper]{article}

\usepackage{cvpr}
\usepackage{times}
\usepackage{epsfig}
\usepackage{graphicx}
\usepackage{amsmath}
\usepackage{amssymb}
\usepackage{bm}
\usepackage{multirow}
\usepackage{array}

\usepackage[breaklinks=true,bookmarks=false]{hyperref}

\cvprfinalcopy 

\def\cvprPaperID{****} 

\begin{document}

\title{An Automated CNN Recommendation System for Image Classification Tasks}

\author{Song Wang, Li Sun, Wei Fan, Jun Sun, Satoshi Naoi, \\
Koichi Shirahata*, Takuya Fukagai*, Yasumoto Tomita*, Atsushi Ike*\\
{Fujitsu Research \& Development Center, Beijing, China}\\
{*Fujitsu Laboratories, Japan}\\
\{wang.song, sun.li, fanwei, sunjun, naoi\}@cn.fujitsu.com\\
\{k.shirahata,  fukagai.takuya, tomita.yasumoto, ike\}@jp.fujitsu.com\\
}

\maketitle

\begin{abstract}
Nowadays the CNN is widely used in practical applications for image classification task. However the design of the CNN model is very professional work and which is very difficult for ordinary users. Besides, even for experts of CNN, to select an optimal model for specific task may still need a lot of time (to train many different models). In order to solve this problem, we proposed an automated CNN recommendation system for image classification task. Our system is able to evaluate the complexity of the classification task and the classification ability of the CNN model precisely. By using the evaluation results, the system can recommend the optimal CNN model and which can match the task perfectly. The recommendation process of the system is very fast since we don't need any model training. The experiment results proved that the evaluation methods are very accurate and reliable.  
\end{abstract}
\section{Introduction}
\label{sec:intro}

The recent development of convolutional neural network (CNN) has made it a powerful tool for image classification tasks. For example, the natural image \cite{szegedy_going_2014} and character image \cite{ciresan_multi-column_2012} classification. This is because the structure of CNN is very suitable for representing various types of images. Now the performance of CNN is good enough for many practical image classification applications. Many companies want to try CNN to ``read'' images instead of people to improve the efficiency and reduce the cost. \par

However the design of CNN needs expert knowledge since there are many parameters to be determined, such as the depth, width and the layer distribution. Moreover, even an expert may still need much time to test different CNN models and then choose the optimal one for the task. If the CNN model is too large for the task, there will be an overfitting problem and waste of computing resource. In contrast, if the model is too small, the accuracy may be much lower than what we can actually achieve (based on the same training data). Consequently, now the urgent demand of the industry is to find a method which can automatically recommend the ``perfect'' CNN model for specific image classification task. This is also helpful for researchers since it can save a lot of time on searching the optimal CNN model. \par 

There are not so many trials which focus on the automated CNN model recommendation. In \cite{mariyama2016automatic} the authors proposed methods which can generate the neural network structure automatically during the training. Nevertheless, these methods can only generate small neural networks (not large deep neural network) for regression problems. To our knowledge, up to now there is no perfect solution to recommend large deep neural network for classification tasks. \par

In this paper, we propose an automated CNN recommendation system for image classification tasks. Different from the trials above, our system is able to recommend very large deep neural network for any image classification task. The detailed merits of the proposed system are shown as follows.
\begin{itemize}
\item By analyzing the training data of a certain classification task, we can directly recommend the optimal CNN model. Compared with the conventional way of model selecting, we don't need to train any model. Therefore, the proposed system can save a lot of time.
\item The analysis of the training data is a quantitative analysis. We evaluate the training data and give a specific ``complexity score'' for it. The complexity score represents the difficulty level of the classification task. As a result, the recommended optimal CNN model based on the complexity score will be very accurate. In other words, our recommended CNN model is able to match the classification task perfectly. The manually selected CNN model may not be that perfectly match since the number of tested model is very limited.
\item The proposed system uses the ``ability score'' to describe the classification ability of the CNN model. The ability score is evaluated by considering the total calculation times, the shape (depth and width) of the model and the gradient vanishing problem. It means that we can give a specific ability score for any CNN model. Consequently, it is able to find the accurate optimal CNN model according to the ability score.
\end{itemize}
\par

This paper is organized as follows. Section \ref{method} introduces the framework of the system and the detailed evaluation method for both classification task (complexity score) and CNN model (ability score). The experimental analysis is shown in Section \ref{experiment}. Finally, Section \ref{conclusion} gives the conclusion and future works to do.\par

\section{methodology}
\label{method}

\begin{figure}
\centerline{\includegraphics[width=0.3\textwidth]{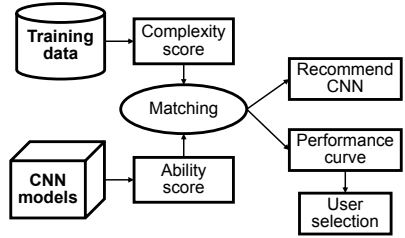}}
\caption{The framework of the proposed system.} \label{framework}
\end{figure}

The framework of the proposed system is shown in Fig.~\ref{framework}. First, the complexity score of the training data of the classification task is evaluated. Meanwhile, the ability scores of the CNN models of different structures are also evaluated. Second, a matching function is created by fitting the ability score to the complexity score. The parameters of the matching function is determined through experiments. Please note that the experiments are only conducted for the matching function estimation. After the matching function is fixed, we don't need to do any experiment for recommending the CNN model. Finally, the system is able to recommend the optimal CNN model for user's classification task. If the user has other requirement, for example, the calculation speed, then the system can generate the performance curve by training the optimal model and an extra model. User is able to choose any model they want according to the curve.\par

\subsection{Complexity score of the training data}

The image classification tasks mainly vary from the following two aspects. First, the class number of the task is various. It may be a number from two to thousands. For example, the MNIST database and the CIFAR10 database both have 10 classes while the ImageNet database has 1,000 classes. Second, the image appearances of different tasks are very different. For example, in character images there are always grayscale strokes with uniform background while in CIFAR or ImageNet databases the images are colorful and has complicated background. Since we have to give the complexity score for all the image classification tasks, both of the variances should be considered in our score evaluation method. \par

\begin{figure}
\centerline{\includegraphics[width=0.45\textwidth]{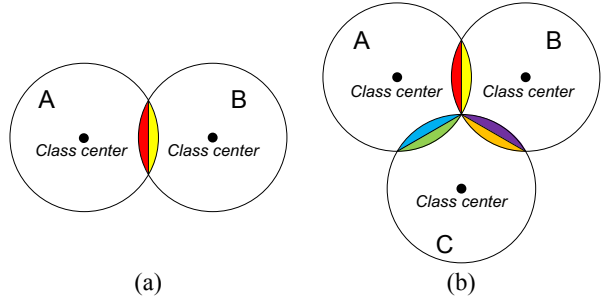}}
\caption{From 2-class problem to more. The large circles stand for the distribution area of class A, B and C.} \label{2-class}
\end{figure}

In the proposed system, the class number variance is solved by converting all the classification tasks into a 2-class problem. This conversion is based on the hypothesis as follows. First of all, we only consider two different classification results: correct or incorrect. That is to say, if the classification result of a test sample is incorrect, we don't care which the incorrect class is. We treat all the incorrect classes as the same. \par

Then assume now we have a classification task of $n$ classes, which are $c_1, c_2, ..., c_n$. If we want to classify a test sample $s \in c_k$, we need some classifier. Assume the classifier can calculate the similarity by function $D()$. If it satisfy the condition
\begin{equation*}
D(s, c_k)>max(D(s,c_i),i \neq k),
\end{equation*}  
the $s$ is correctly classified. This condition means that for $s$, if its similarity to $c_k$ is larger than the specific class $c_m (m=\arg \underset{i}{max}D(s, c_i)$), the $s$ is correctly classified. In this classification process, except $c_k$ and $c_m$, the rest of the classes will not influence the classification result. In other words, the classification result of $s$ is determined by the comparison between its similarities to $c_k$ and $c_m$. Consequently, our hypothesis is that, for this test sample $s$, its classification can be seen as a 2-class classification problem of $c_k$ (the test sample belongs to) and $c_m$ (the class with the highest similarity from the classes that the test sample doesn't belong to). \par

The above hypothesis can also be interpreted in the following way. The classification process of $s$ can be seen as a group of independent 2-class classification problems: $(c_k, c_i), i \in (1,n), i \neq k$. If the classification results of these 2-class problems are all correct, the final classification result of $s$ is correct. Otherwise, the result is incorrect. \par

For example, as shown in Fig.~\ref{2-class} (a), for an original 2-class ($A$ and $B$) problem, if we simply use the distance to the class center as the similarity measurement, we can divide the sample distribution area into several parts according to the classification result. The $A$ class samples in yellow area and the $B$ class samples in red area will be classified incorrectly since they are closer to the incorrect class center. Assume all the samples are evenly distributed in each class, then for each class, the classification error rate $e$ is the ratio of the colored area of incorrectly classification to the circle of the class. Obviously, the error rate of this 2-class problem is also $e$. \par

As shown in Fig.~\ref{2-class} (b), if we add a new class $C$, then the 2-class problem becomes 3-class problem. Assume there is no overlapping among all the colored area, then for each class, the error rate is doubled since the color area of incorrectly classification is doubled. As a result, the error rate of this 3-class problem is $2e$. This means that if we ignore the overlapping of the colored area, the error rate of a classification task is proportional to its class number. In conclusion, it is reasonable to describe the complexity of a classification task by converting it to a 2-class problem. \par

After converting the classification task into 2-class problem, the complexity score is given as follows. We define that the complexity score should describe the complexity for a certain classifier to classify the training data. Therefore, each sample in the training data should have a complexity score to describe the difficulty for the classifier to classify it. The complexity score $C$ for sample $s$ ($s \in c_k$) is
\begin{equation}
C = \dfrac{D(s, c_k)}{D(s, c_k)+max(D(s,c_i),i \neq k)}.
\label{cs}
\end{equation} 
The complexity score $C_{all}$ for the whole training data is just the average score of all the training samples. Assume the total sample number of the training data is $l$, then $C_{all}$ is calculated as
\begin{equation}
C_{all}=\overline{C}=\dfrac{1}{l} \sum^{l}_{j=1}C_j.
\label{csf}
\end{equation}  

\begin{figure}
\centerline{\includegraphics[width=0.35\textwidth]{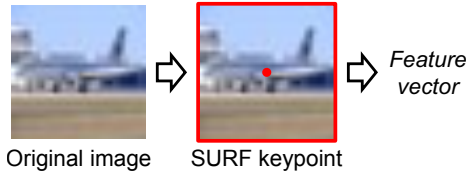}}
\caption{SURF feature vector extraction. The red bounding box represents the SURF keypoint.} \label{SURF}
\end{figure}

In order to solve the image variance of different tasks, a simple classifier and general feature extraction method are used for calculating the complexity score. First, as shown in Fig.~\ref{SURF}, for each training sample, a SURF~\cite{bay_surf:_2006} descriptor is extracted by considering the whole image as one single keypoint. The size of the keypoint is the same with the image and the center of which is just the image center. Besides, the rotation angle of this keypoint is also set to $0$. By doing so, each training sample is represented as a feature vector (sample vector). Since SURF is a manually designed feature for general image processing tasks, it has been widely used for various image classification tasks~\cite{khan2011sift,uchida_part-based_2010}. Therefore, the SURF feature can represent the variance image appearances in a stable way. Second, a simple centroid classifier is employed to classify the training samples in the feature space. Similarly, for the stability, this classifier is not obtained by learning from samples. For each class, the centroid is just the average of all the sample vectors of this class. The similarity is just measured by the Euclidean distance between the sample vector and the centroid. Assume $\bm{\vec s}$ is a certain sample vector and $\bm{\vec c}$ is some centroid, then the similarity of this sample vector to this centroid is given as
\begin{equation*}
D(\bm{\vec s},\bm{\vec c})=e^{-\left|\left|\bm{\vec s}-\bm{\vec c}\right|\right|}.
\end{equation*}  
With this similarity calculation function, according to \eqref{cs} and \eqref{csf}, the complexity score can be calculated.  \par

\subsection{Ability score for CNN models}

There are a lot of parameters which can determine the structure of the CNN model. Therefore, in the proposed system, we first give a CNN model generation method and then evaluate the ability score of each generated CNN model by considering the parameters. In order to reduce the variance, in the generation method, several parameters are fixed. The unfixed parameters are also changed under strict rules.\par

The fixed parameters are shown as follows. These parameters have weak influence on the classification ability of the CNN model, thus we can directly use optimal settings.
\begin{itemize}
\item \textbf{Input image size:} $52\times52$.
\item \textbf{Convolutional kernel size:} $3\times3$ for all the convolutional layers.
\item \textbf{Convolutional kernel stride:} $1$ or $2$, depends on whether we need shrinking the feature map size or not.
\item \textbf{Bounding box padding size:} $1$, since the kernel size is $3\times3$, this size can ensure that the feature map size remains the same after the convolutional calculation (if stride is $1$). 
\item \textbf{Pooling kernel size:} $3\times3$ for all the pooling layers.
\item \textbf{Pooling kernel stride:} $2$, this means after the pooling layer the feature map size will be reduced by half. 
\item \textbf{Pooling layer type:} max pooling.
\item \textbf{Activation function:} ReLU.
\end{itemize}
Besides these parameters of the CNN model, several settings of the model training are also fixed, such as the learning rate and training optimizing methods (dropout~\cite{Srivastava2014} and batch normalization~\cite{ioffe2015batch}). \par
\begin{figure}
\centerline{\includegraphics[width=0.4\textwidth]{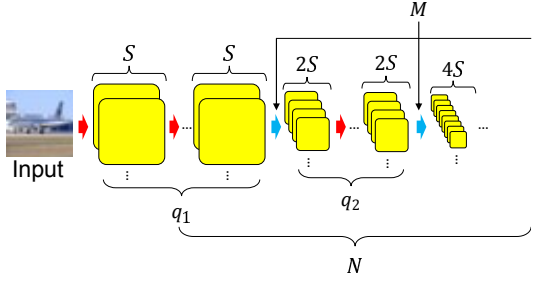}}
\caption{The CNN model generation rules. The yellow squares stand for feature maps. The red arrow shows the convolutional layer while the blue arrow shows the down-sampling layer.} \label{CNN}
\end{figure}

The ability score of the generated CNN model are mainly determined by the unfixed parameters. As shown in Fig.~\ref{CNN}, the unfixed parameters are described as follows.
\begin{itemize}
\item \textbf{Convolutional layer number:} denoted as $N$. The depth of the CNN model is mainly determined by $N$.
\item \textbf{Down-sampling layer number:} denoted as $M$. The down-sampling layer means that after this layer the number of the feature maps is doubled while the size of which is reduced by half. The down-sampling layer can be pooling layer or convolutional layer with stride of 2. Between two neighbor down-sampling layers (or between the input image layer and the first down-sampling layer), the numbers and sizes of feature map of different convolutional layers are the same. Since the number of convolutional layers between different down-sampling layers may be different, the $\bm{\vec q}=q_1, q_2, ..., q_M$ are used to denote these numbers of different section. Obviously, we have $ N=\sum^{M}_{i=1}q_i $.
\item \textbf{Feature map numbers:} which are changed according to the rule described above. Therefore, let $S$ denote the feature map number of the first convolutional layer, then the numbers of feature map of the rest convolutional layers can be determined. Consequently, $S$ can be seen as the basic feature map number and by which the width of the model is determined.
\end{itemize}

After the model generation, the next step is to calculate the ability score. Besides the parameters of the structure of the CNN model, the gradient vanishing problem of training is also considered. Let $\chi$ denotes the ability score and which is calculated by the function as
\begin{equation}
\chi = f(N,S,M, \bm{\vec q})g(N).
\label{as}
\end{equation} 
In \eqref{as}, the $f()$ gives the score according to the structure of the CNN model. If the structure is larger, the $f()$ is higher. The measurement of the structure is mainly based on the total number of calculations of the model. The $g()$ is a correction based on the gradient vanishing problem, thus it is related to the model depth $N$. With $g()$, the $\chi$ will be decreased if $N$ is too large. Several generated CNN models and corresponding ability score are shown in Table.~\ref{models}. These models will be used for the experimental analysis. \par

\begin{table}[ht]
\small
\caption{Generated CNN models.} 
\vspace{0.5\baselineskip}
\centering
{\begin{tabular}{c|c|c|c|c|c } 
Model 				& $N$ & $S$ & $M$ & $\bm{\vec q}$ & $\chi$ \\ \hline
Model-1 			& 3 & 16 & 3 & $(1,1,1)$ & 5.41 \\
Model-2 			& 4 & 16 & 4 & $(1,1,1,1)$ & 5.44 \\
\textbf{Model-3} 	& 4 & 64 & 4 & $(1,1,1,1)$ & 6.04 \\
Model-4 			& 6 & 64 & 4 & $(1,1,2,2)$ & 6.12 \\
\textbf{Model-5}	& 8 & 64 & 4 & $(2,2,2,2)$ & 6.34 \\
Model-6				& 13 & 64 & 4 & $(3,3,3,4)$ & 6.53 \\
\end{tabular}
}
\label{models}
\end{table}

One interesting observation is that, because of $g()$, the $\chi$ has a maximum value. This is because compared with the depth, the width of the CNN model has very little contribution to its classification ability. In other words, no matter how deep is the CNN model, its classification ability is limited to a certain value. Consequently, although now we have very powerful hardware for CNN, the ability of which is limited. \par

\subsection{Matching function}

After the complexity score of the database and the ability score of the CNN models are obtained, the next step is to find the relationship between these two scores. Let $m()$ denotes the matching function and which is defined as
\begin{equation}
\chi = m(C_{all}).
\label{match}
\end{equation} 
In which the $\chi$ is the ability score and $ C_{all}$ is the complexity score. This matching function means that for an input complexity score of a certain classification task we can find its corresponding ability score. Afterwards the CNN model with the same ability score is the matching model for this task. \par

The matching function is obtained by testing several classification tasks. In our definition, if one CNN model can just achieve 100\% classification rate on training data of a certain task, it is seen as the optimal model for this task. This is because theoretically, for a CNN model, it can only learn from the training data for classification. When a CNN model has higher ability than achieving 100\% classification rate on training data, there will be a risk of overfitting. Therefore, for a test classification task, we will match it with the ability score of the CNN model which can just achieve 100\% classification rate on training data. With several test tasks and their matching ability score, we can build the matching function $m()$.\par
 
However, there is also some experience which indicates that the model of the best performance is always a little bit overfitting for the task. Consequently, we will recommend those CNN models which has a little bit higher ability score than the optimal model for the classification task. \par 

\begin{figure*}
\centerline{\includegraphics[width=0.8\textwidth]{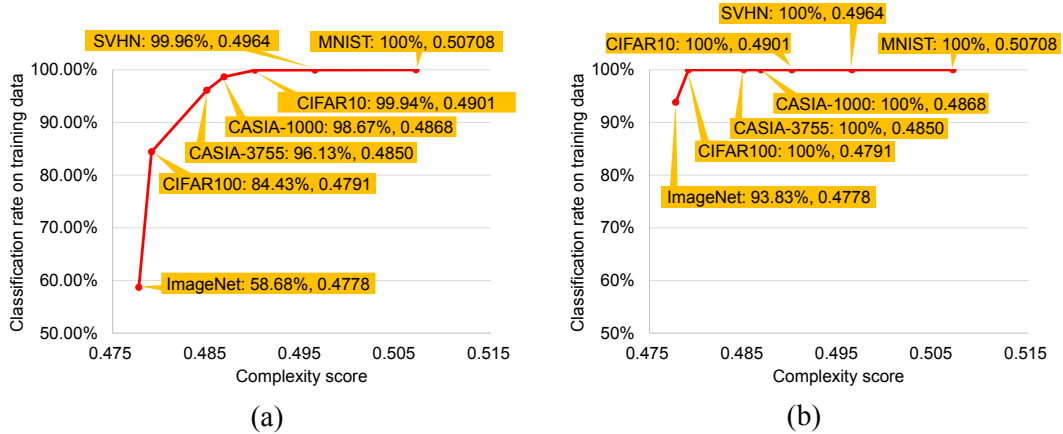}}
\caption{The reliability of complexity score. (a) shows the experimental results on small model (Model-3) while (b) shows the experimental results on large model (Model-5). } \label{model3}
\end{figure*}

\section{Experimental analysis}
\label{experiment}

The experiments were divided into three parts: the experiments of complexity score, the experiments of ability score and the experiments of the recommendation. The experiments of first two parts were used to test the reliability of the evaluated complexity score and ability score. The experiments of last part were used to show the process of the whole system. 

\subsection{Experiments of complexity score}

In order to test the reliability of our evaluation method for complexity score, a variety of image classification databases were used. These databases are shown as follows.
\begin{itemize}
\item MNIST: an image database of handwritten digits. The class number is 10. This database contains 60,000 training samples and 10,000 test samples.
\item SVHN: the google street view images of house numbers. The class number is 10. This database contains 73,257 training samples and 26,032 test samples.
\item CASIA-3755: an database of handwritten characters from CASIA-HWDB. We used its competition version~\cite{yin_icdar_2013} and which contains 3,755 classes. This database contains 2,148,324 training samples and 534,563 test samples.
\item CASIA-1000: the same with the database introduced above. This database can be seen as a subset and which is simply created by selecting 1,000 classes from CASIA-3755. This database contains 573,302 training samples and 141,389 test samples.
\item CIFAR10: a database of images of different objects. The class number is 10. This database contains 50,000 training samples and 10,000 test samples.
\item CIFAR100: a database just like CIFAR10, except it contains 100 classes. There are 50,000 training samples and 10,000 test samples. 
\item ImageNet: a very large image database and which contains variety of objects, scenes and so on. The class number is 1,000. There are 1,281,167 training samples and 50,000 test samples.
\end{itemize}

The databases listed above were tested on two different CNN models: Model-3 and Model-5 of Table.~\ref{models}. Since the same model has the same ability score, if we test different databases on the same model, we should find a positive correlation between their classification rates of training data and their complexity scores. The test results of Model-3 are shown in Fig.~\ref{model3} (a). Obviously, with the increase of the complexity score, the classification rate got higher and higher. This proved that the complexity score is reliable for describing the complexity of the task. In Fig.~\ref{model3} (b), since Model-5 is larger than Model-3, most of the databases achieved 100\% classification rate except the ImageNet.

\subsection{Experiments of ability score}

\begin{figure}
\centerline{\includegraphics[width=0.4\textwidth]{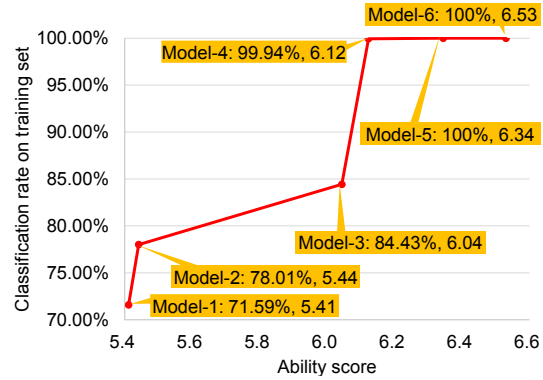}}
\caption{The reliability of ability score. The experiments were conducted on CIFAR100} \label{abilitys}
\end{figure}

Similar with the experiments of complexity score, in order to evaluate the reliability of the ability score, different CNN models were tested on the same database. As shown in Fig.~\ref{abilitys}, on the CIFAR100, all the CNN models of Table.~\ref{models} were tested. Similarly, according to the experimental results, it is able to find a positive correlation between the classification rates of training data and ability score. However, the curve was not as smooth as the curve of the complexity score. This means that the we still need to improve the ability score evaluation method. \par

\subsection{Recommendation by performance curve}

Besides directly recommending the optimal CNN, the system is also able to generate the performance curve for users to select any CNN model they need. The performance curve describes the relationship between the classification rate on validation data and the average forward processing time of the models. With this curve, users can find balance models between accuracy and speed according to their own demand. \par

\begin{figure}
\centerline{\includegraphics[width=0.47\textwidth]{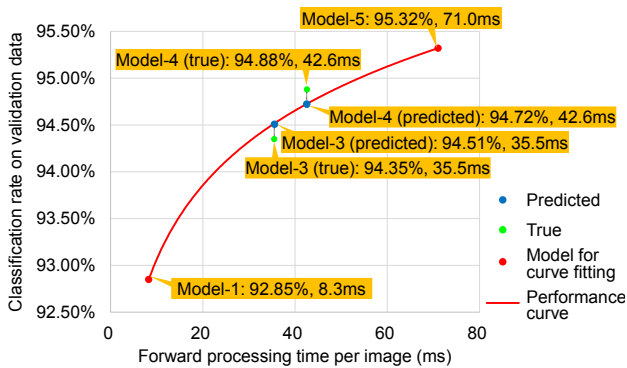}}
\caption{The performance curve of SVHN.} \label{curve}
\end{figure}

In order to obtain the performance curve, we have to train two CNN models and get their classification rate on validation data: one is the recommended optimal CNN model and the other is a smaller model. The classification rate on validation data can represent the performance of the model in practical application. In the experiments, the SVHN was used as the test task and the rest of the databases were used for obtaining the matching function \eqref{match}. The performance curve of SVHN is shown in~Fig.\ref{curve}. Model-5 of Table.~\ref{models} was the recommended optimal model for SVHN and Model-1 was the extra model for the performance curve fitting. Model-3 and Model-4 were tested as the user selected model. The true classification rate of Model-3 and Model-4 were obtained by experiments. Clearly, the predicted classification rate of Model-3 and Model-4 were very close to their true value. This proved that the performance curve can be an reliable reference for user to select CNN models. \par

\section{Conclusion}
\label{conclusion}

In this paper, we proposed an system which can evaluate the complexity score of the classification task as well as the classification ability score of the CNN model. By using these two scores, it is possible to recommend proper CNN model for users who are not familiar with deep learning technology. Besides, the recommendation process doesn't need any training process of the CNN model, thus it is very fast. 

In the future, we will continue to improve the accuracy of the evaluation of the two scores. Besides, it is possible to extend the proposed system to other tasks of deep learning, for example, the object detection. \par

{\small
\bibliographystyle{ieee}
\bibliography{References}

\begin{thebibliography}{1}\itemsep=-1pt

\bibitem{bay_surf:_2006}
H.~Bay, T.~Tuytelaars, and L.~Van~Gool.
\newblock Surf: {Speeded} up robust features.
\newblock In {\em Computer vision\textemdash ECCV 2006}, pages 404--417.
  Springer, 2006.

\bibitem{ciresan_multi-column_2012}
D.~Ciresan, U.~Meier, and J.~Schmidhuber.
\newblock Multi-column deep neural networks for image classification.
\newblock In {\em 2012 {IEEE} {Conference} on Computer {Vision} and {Pattern}
  {Recognition}}, pages 3642--3649. IEEE, 2012.

\bibitem{ioffe2015batch}
S.~Ioffe and C.~Szegedy.
\newblock Batch normalization: Accelerating deep network training by reducing
  internal covariate shift.
\newblock {\em arXiv preprint arXiv:1502.03167}, 2015.

\bibitem{khan2011sift}
N.~Y. Khan, B.~McCane, and G.~Wyvill.
\newblock Sift and surf performance evaluation against various image
  deformations on benchmark dataset.
\newblock In {\em Digital Image Computing Techniques and Applications (DICTA),
  2011 International Conference on}, pages 501--506. IEEE, 2011.

\bibitem{mariyama2016automatic}
T.~Mariyama, K.~Fukushima, and W.~Matsumoto.
\newblock Automatic design of neural network structures using ais.
\newblock In {\em International Conference on Neural Information Processing},
  pages 280--287. Springer, 2016.

\bibitem{Srivastava2014}
N.~Srivastava, G.~Hinton, A.~Krizhevsky, I.~Sutskever, and R.~Salakhutdinov.
\newblock Dropout: {A} {Simple} {Way} to {Prevent} {Neural} {Networks} from
  {Overfitting}.
\newblock {\em J. Mach. Learn. Res.}, 15(1):1929--1958, Jan. 2014.

\bibitem{szegedy_going_2014}
C.~Szegedy, W.~Liu, Y.~Jia, P.~Sermanet, S.~Reed, D.~Anguelov, D.~Erhan,
  V.~Vanhoucke, and A.~Rabinovich.
\newblock Going deeper with convolutions.
\newblock {\em arXiv preprint arXiv:1409.4842}, 2014.

\bibitem{uchida_part-based_2010}
S.~Uchida and M.~Liwicki.
\newblock Part-based recognition of handwritten characters.
\newblock In {\em 2010 {International} {Conference} on Frontiers in
  {Handwriting} {Recognition} ({ICFHR}),}, pages 545--550. IEEE, 2010.

\bibitem{yin_icdar_2013}
F.~Yin, Q.-F. Wang, X.-Y. Zhang, and C.-L. Liu.
\newblock Icdar 2013 chinese handwriting recognition competition.
\newblock In {\em 2013 12th {International} {Conference} on Document {Analysis}
  and {Recognition} ({ICDAR})}, pages 1464--1470. IEEE, 2013.

\end{thebibliography}
}

\end{document}